\pdfoutput=1

\documentclass[11pt]{article}

\usepackage[final]{acl}

\usepackage{times}
\usepackage{latexsym}

\usepackage[T1]{fontenc}

\usepackage[utf8]{inputenc}

\usepackage{microtype}

\usepackage{inconsolata}

\usepackage{booktabs}
\usepackage{graphicx}

\title{Leveraging Transformer-Based Models for Predicting Inflection Classes of Words in an Endangered Sami Language}

\author{Khalid Alnajjar \\
  Rootroo Ltd \\
  \texttt{first@rootroo.com} \\\And
  Mika Hämäläinen \\
  Metropolia University\\of Applied Sciences \\
  \texttt{first.last@metropolia.fi} \\\And
  Jack Rueter \\
  University of Helsinki \\
  \texttt{first.last@helsinki.fi} \\}

\begin{document}
\maketitle
\begin{abstract}

This paper presents a methodology for training a transformer-based model to classify lexical and morphosyntactic features of Skolt Sami, an endangered Uralic language characterized by complex morphology. The goal of our approach is to create an effective system for understanding and analyzing Skolt Sami, given the limited data availability and linguistic intricacies inherent to the language. Our end-to-end pipeline includes data extraction, augmentation, and training a transformer-based model capable of predicting inflection classes. The motivation behind this work is to support language preservation and revitalization efforts for minority languages like Skolt Sami. Accurate classification not only helps improve the state of Finite-State Transducers (FSTs) by providing greater lexical coverage but also contributes to systematic linguistic documentation for researchers working with newly discovered words from literature and native speakers. Our model achieves an average weighted F1 score of 1.00 for POS classification and 0.81 for inflection class classification. The trained model and code will be released publicly to facilitate future research in endangered NLP.

\end{abstract}

\section{Introduction}

Skolt Sami is a minority language in the Uralic family, spoken primarily in Finland, and is characterized by complex morphosyntactic properties and rich morphological forms (see \citealt{feist2011grammar}). Minority languages like Skolt Sami face significant challenges in the field of natural language processing (NLP) due to their endangered nature, including a lack of extensive annotated datasets and linguistic resources. This scarcity complicates the development of computational models capable of effectively understanding and analyzing the language. Moreover, the morphology of Skolt Sami is highly intricate, with numerous inflections and derivations that present additional challenges for automated processing \cite{rueter2020fst}. Despite these challenges, developing NLP models for minority languages is essential to preserve linguistic diversity and support language revitalization.

Accurate part-of-speech (POS) and inflection class classification are fundamental steps in understanding the grammatical and semantic structure of a language. Such classifications enable downstream NLP applications like machine translation, morphological analysis, and syntactic parsing, which are particularly important for languages with rich morphology. Additionally, effective classifiers can assist in improving the current state of FSTs by providing greater lexical coverage, ultimately enhancing their ability to handle the full range of morphological variations found in Skolt Sami. Classifiers can also aid researchers in systematically documenting new words collected from literature and native speakers, which is crucial for tracking linguistic evolution in endangered contexts. For Skolt Sami, POS and inflection class classification can contribute to building digital resources and educational tools, making the language more accessible to both linguists and speakers.

To address these challenges, we propose a transformer-based model designed to automate the analysis of Skolt Sami, specifically for POS and inflection class classification. Our approach includes data extraction, preprocessing, augmentation, model training and evaluation. We employed advanced transformer architectures to learn the linguistic features of Skolt Sami effectively. Additionally, we provide both the trained model and the accompanying code publicly to support future research efforts on endangered languages\footnote{\href{https://github.com/mokha/predict-inflection-class}{https://github.com/mokha/predict-inflection-class}}.

The contributions of this work are as follows:

\begin{enumerate}
    \item \textbf{Data Augmentation Using Miniparadigms}: We employed data augmentation techniques, including the generation of morphological forms, to mitigate data scarcity and improve model robustness.
    \item \textbf{Transformer-Based Model}: We designed a transformer-based model for POS and inflection class classification in Skolt Sami, utilizing shared embedding layers and task-specific output heads.
\end{enumerate}

\section{Related work}

Skolt Sami has received a moderate amount of NLP research interest as a result of Dr Jack Rueter's amazing work on building the fundamental NLP building blocks for Skolt Sami\footnote{https://researchportal.helsinki.fi/en/projects/ koltansaamen-elvytys-kieliteknologia-avusteisen-kielenoppimisohje} as a result, Skolt Sami has an FST \cite{rueter2020fst} an online dictionary (see \citealt{hamalainen2021online}), a Universal Dependencies treebank \cite{nivre2022ud_skolt_sami} and some neural models to identify cognates \cite{hamalainen2019finding}.

An empirical study by \citealt{wu2020applying} reveals that the transformer’s performance on character-level transduction tasks, such as morphological inflection generation, is significantly influenced by batch size, unlike in recurrent models. By optimizing batch size and introducing feature-guided transduction techniques, the transformer can outperform RNN-based models, achieving state-of-the-art results on tasks such as grapheme-to-phoneme conversion, transliteration, and morphological inflection. This study demonstrates that, with appropriate modifications, transformers are highly effective for character-level tasks as well.

Recent research \cite{abudouwaili-etal-2023-joint} has introduced a joint morphological tagger specifically designed for low-resource agglutinative languages. By leveraging multi-dimensional contextual features of agglutinative words and employing joint training, the proposed model mitigates the error propagation typically seen in part-of-speech tagging while enhancing the interaction between part-of-speech and morphological labels. Furthermore, the model predicts part-of-speech and morphological features separately, using a graph convolution network to capture higher-order label interactions. Experimental results demonstrate that this approach outperforms existing models, showcasing its effectiveness in low-resource language settings.

One notable contribution in this area is a transformer-based inflection system that enhances the standard transformer architecture by incorporating reverse positional encoding and type embeddings proposed by \citet{yang-etal-2022-generalizing}. To address data scarcity, the model also leverages data augmentation techniques such as data hallucination and lemma copying. The training process is conducted in two stages: initial training on augmented data using standard backpropagation and teacher forcing, followed by further training with a modified version of scheduled sampling, termed student forcing. Experimental results demonstrate that this system achieves competitive performance across both small and large data settings, highlighting its efficacy in diverse morphological inflection tasks.

Recent work \cite{hamalainen-etal-2021-neural} on morphological analysis, generation, and lemmatization for morphologically rich languages has focused on training recurrent neural network (RNN)-based models. A notable contribution in this area is the development of a method for automatically extracting large amounts of training data from finite-state transducers (FSTs) for 22 languages, including 17 endangered ones. These neural models are designed to follow the same tagset as the FSTs, ensuring compatibility and allowing the neural models to serve as fallback systems when used in conjunction with the FSTs. This approach enhances the accessibility and preservation of endangered languages by leveraging both neural and rule-based systems.

\section{Methodology}

\subsection{Data Collection and Preparation}

Data extraction and preprocessing are particularly critical when working with an endangered language like Skolt Sami. This phase involved extracting linguistic data from available resources and transforming it into a structured format suitable for further processing.

We extracted a total of 28,984 lexemes from Ve\textsuperscript{$\prime$}rdd \cite{alnajjar2020ve}, an online tool designed for editing and managing dictionaries for endangered languages. Ve\textsuperscript{$\prime$}rdd offers a structured and efficient way to curate linguistic resources, making it an invaluable asset for our dataset creation process. The extracted lexemes included diverse entries from the dictionary, which were parsed and transformed into a tabular format for further analysis and training. This structured dataset stored each lexeme along with its POS and contextual lexical information, ensuring consistency and accessibility for subsequent processing.

\subsection{Data Cleaning and Filtering}

Data cleaning and filtering are crucial in the context of endangered languages to ensure data quality and improve model performance. We filtered the dataset to include only nouns (\verb|N|) and verbs (\verb|V|), as these categories were the most frequent and useful for subsequent morphological analysis. These POS categories were selected due to their high occurrence and significance in understanding the linguistic structure of Skolt Sami.

We further filtered lexemes based on specific patterns using regular expressions, removing non-standard or infrequent forms to enhance the model's ability to generalize to common usage patterns.

\subsection{Data Augmentation Using Miniparadigms}

To mitigate data scarcity, we employed data augmentation using "miniparadigms." For each verb and noun, specific morphological forms (e.g., present tense, singular form, imperative) were generated. We have employed UralicNLP \cite{hamalainen2019uralicnlp} with PyHFST \cite{alnajjar2023pyhfst} as the backend and used Skolt Sami FST transducer \cite{rueter2020fst} to generate the forms. This approach added multiple derived forms for each lexeme, thereby significantly increasing the size of the dataset. The use of miniparadigms allowed the model to learn morphological variations more effectively, compensating for the limited data available.

Table~\ref{tab:selected-forms} lists the miniparadigms used for data augmentation. These generated forms helped increase the robustness and generalization capability of the model.

\begin{table*}[]
\resizebox{\textwidth}{!}{%
\begin{tabular}{@{}ll@{}}
\toprule
\textbf{POS} & \textbf{Morphological Forms Generated}                                                                                                                                                          \\ \midrule
V (Verbs)    & \begin{tabular}[c]{@{}l@{}}V+Ind+Prs+ConNeg, V+Ind+Prs+Sg3, V+Ind+Prt+Sg1, V+Ind+Prt+Sg3, \\ V+Inf, V+Ind+Prs+Sg1, V+Pass+PrfPrc, V+Ind+Prs+Pl3, V+Imprt+Sg3, V+Imprt+Pl3\end{tabular} \\\hline
N (Nouns)    & \begin{tabular}[c]{@{}l@{}}N+Sg+Loc, N+Sg+Ill, N+Pl+Gen, N+Sg+Nom, N+Sg+Gen, N+Sg+Loc+PxSg3, \\ N+Ess, N+Der/Dimin+N+Sg+Nom, N+Der/Dimin+N+Sg+Gen, N+Sg+Ill+PxSg1\end{tabular}         \\ \bottomrule
\end{tabular}%
}
\caption{Select morphological forms to be used in the data augmentation phase}
\label{tab:selected-forms}
\end{table*}

\subsection{Contlex Cleaning and Filtering}

In total, there were 939 unique continuation lexica (Contlex) for nouns (N) and verbs (V). Contlexes are an FST way of indicating that a word belongs to a certain inflection class. Many of these Contlex labels included additional information, such as \verb|V_JOAQTTED_ERRORTH|. To standardize the dataset, we removed any additional information following the second underscore (\verb|_|). This process reduced the number of unique Contlex labels to 514.

However, a large portion of these Contlex categories had very few lexemes. To improve data quality and model robustness, we filtered out any Contlex category that had fewer than 50 lexemes as part of the data cleaning phase. After this filtering, we ended up with 73 Contlex categories — 52 for nouns and 21 for verbs. Table~\ref{tab:supported-contlex} lists the supported Contlex for each part-of-speech.

\begin{table}[]
\resizebox{0.83\columnwidth}{!}{%
\begin{tabular}{@{}cl@{}}
\toprule
\textbf{POS} & \textbf{Contlex Supported}                                                                                                                                                                                                                                                                                                                                                                                                                                                                                                   \\ \midrule
N            & \begin{tabular}[c]{@{}l@{}}SAAQMM, SAJOS, MAINSTUMMUSH, \\ AELDD, CHAAQCC, VUYRR, ALGG, \\ TUYJJ, CHUOSHKK, CHUAQRVV, \\ KAADHNEKH, MUORYZH, TAQHTT, \\ PAPP, JEAQRMM, AANAR, MUORR, \\ VOONYS, TAALKYS, AUTT, LOAQDD, \\ BIOLOGIA, PAIQKHKH, KUEQLL, \\ PIEAQSS, KAQLBB, PLAAN, NEAVVV, \\ JAEUQRR, PAARR, PESS, JUQVJJ, \\ PEAELDD, HOQPPI, KUEAQTT, \\ KUYLAZH, MIYRKK, MEERSAZH, \\ AACCIKH, TOLL, JEAQNNN, ATOM, \\ JUURD, PEIQVV, SIJDD, KHEQRJJ, \\ MIEAQRR, MUEQRJJ, PAAQJJ, \\ SIYKKK, SHOOMM, OOUMAZH\end{tabular} \\ \hline
V            & \begin{tabular}[c]{@{}l@{}}LAUKKOOLLYD, SILTTEED, \\ TEEQMEED, ILAUKKOOLLYD, \\ VOQLLJED, KAEQTTED, SOLLEED, \\ KHIORGGNED, SARNNAD, AALGXTED, \\ SHORRNED, KUYDHDHDHJED, \\ KHEEQRJTED, TVOQLLJED, VIIKKYD, \\ JEAELSTED, CEQPCCED, POOLLYD, \\ SHKUEAQTTED, TOBDDYD, ROVVYD\end{tabular}                                                                                                                                                                                                                                   \\ \bottomrule
\end{tabular}%
}
\caption{List of supported Contlex for each POS}
\label{tab:supported-contlex}
\end{table}

\subsection{Tokenization}

To handle the morphological complexity of Skolt Sami, we employed Byte-Pair Encoding (BPE) \cite{gage1994new} as a tokenization method. BPE is particularly effective for morphologically rich languages as it provides subword tokenization that allows the model to understand both frequent morphemes and unique words. We trained a BPE model on the concatenated lexeme and all the form data generated, using a vocabulary size of 2000 to capture the most relevant subword units for the language.

This tokenization approach helped the model deal with highly inflected forms of lexemes by breaking them into smaller, more manageable units, allowing for improved learning over the entire lexicon. The tokenized output was then integrated back into the dataset for model training.

\subsection{Label Encoding}

The dataset involved categorical features such as parts of speech and contextual lexical categories, which needed to be converted into numerical form. We designed a custom label encoder that used one encoder for parts of speech and a separate encoder for each POS-specific lexical category. This hierarchical encoding strategy preserved the information about POS categories while ensuring flexibility for lexical predictions.

The encoded labels were split into training and testing sets, ensuring stratified sampling was used to maintain the distribution of labels, especially given the limited dataset size.

\subsection{Transformer Model Architecture}

We designed a transformer-based \cite{vaswani2017attention} neural network where we employed a shared embedding layer followed by a transformer encoder to learn generalized representations for both tasks: POS prediction and Contlex prediction. The model architecture involved a sequence of well-justified choices aimed at optimizing learning while maintaining simplicity and efficiency.

The input tokens, which were first processed using Byte-Pair Encoding, were then passed through a shared embedding layer. This embedding layer learned a consistent representation for all input data, regardless of the specific task. We opted for a shared embedding layer to leverage common linguistic features across POS and Contlex prediction tasks, ensuring that the model's parameters were efficiently utilized. By sharing these embeddings, we aimed to capture general patterns in Skolt Sami morphology that were common to both POS tagging and inflection class categorization.

The transformer encoder consisted of two encoder layers with four attention heads each. This configuration was chosen to balance the need for model depth and computational efficiency. The attention mechanism allowed the model to capture dependencies between tokens effectively, which is crucial for understanding the morphosyntactic structure of Skolt Sami. The use of multiple attention heads enabled the model to focus on different aspects of token relationships, allowing for a more nuanced understanding of linguistic features.

At the end of the architecture, we implemented separate output heads for each classification task—one for POS classification and one for Contlex classification. These output heads ensured that the model optimized separately for each task, while still sharing the underlying representations learned through the shared embedding and transformer layers. This approach allowed the model to benefit from multi-task learning, where the training process for one task could enhance learning for the other due to shared morphological features.

We have applied the Xavier uniform distribution~\cite{pmlr-v9-glorot10a} on the embeddings and classification layers to initialize the weights, this is to ensure that the variance of the activations stays consistent across layers, which is particularly important in deep networks like transformers to prevent vanishing or exploding gradients during training.

\subsection{Training}

We employed the following training strategies to improve the model's performance and optimize resource usage. The transformer model was trained with a consistent set of hyperparameters throughout the experiments. The embedding size was set to 128, the hidden layer size to 512, and a learning rate of 0.003 was used. A batch size of 512 ensured that the training was efficient while reducing overfitting risk. Hyperparameter optimization was conducted using grid search to identify the optimal settings for dropout rates, the number of layers, and the type of learning rate scheduler.

We have employed AdamW optimizer~\cite{loshchilov2017decoupled} because it combines the benefits of adaptive learning rates with weight decay, which helps in better generalization by decoupling the weight decay from the learning rate schedule. Moreover, we experiment with different schedulers, namely Cosine Annealing, which gradually decreases the learning rate following a cosine curve to allow for fine-tuning near the end of training~\cite{loshchilov2016sgdr}, Exponential, which reduces the learning rate by a fixed factor after every epoch for steady decay~\cite{Li2020An}, and ReduceLROnPlateau, which lowers the learning rate when the performance of the model stops improving.

The model was trained for 100 epochs without early stopping. At epoch 80, the learning rate scheduler was replaced with `SWALR` (Stochastic Weight Averaging Learning Rate) to further refine the model parameters during the final phase of training. SWA has been demonstrated to improve model generalization by allowing the model to converge to a wider minimum in the loss landscape~\cite{izmailov2018averaging}. This approach helps reduce overfitting and often results in better generalization on the test set, particularly for complex neural architectures like transformers.

We did not use mixed-precision training; instead, we kept the precision consistent throughout the experiments to ensure model stability and reproducibility. During training, checkpoints were periodically saved based on the validation metrics to ensure the optimal version of the model was retained for further evaluation.

The loss function combined cross-entropy losses from both POS and Contlex output heads, with adjustable weights for each loss to balance the importance of both tasks. We gave both an equal weight of 1.0. This multi-task learning approach allowed the model to leverage shared morphological and syntactic information while optimizing for distinct objectives.

\section{Results}

We conducted six different training experiments to determine the optimal hyperparameter settings for POS and Contlex classification. The batch size, embedding size, and hidden layer size were consistent across all experiments, set to 512, 128, and 512 respectively. The following table summarizes the different setups and their corresponding performance metrics for both tasks:

\begin{table*}[]
\resizebox{\textwidth}{!}{%
\begin{tabular}{@{}lllllll@{}}
\toprule
Experiment ID & Scheduler Type                 & Dropout & N\_layers & N\_heads & POS F-1 Score & Contlex F-1 \\ \midrule
Exp 1         & CosineAnnealingLR, T\_max=25   & 0.1     & 2          & 4         & 0.93         & 0.64             \\\hline
Exp 2         & CosineAnnealingLR, T\_max=25   & 0.2     & 3          & 4         & 1.00         & 0.78             \\\hline
Exp 3         & CosineAnnealingLR, T\_max=25   & 0.2     & 3          & 8         & \textbf{1.00}         & \textbf{0.81}             \\\hline
Exp 4         & ExponentialLR, gamma=0.95      & 0.2     & 3          & 8         & 0.96         & 0.75             \\\hline
Exp 5         & ReduceLROnPlateau, patience=10 & 0.2     & 3          & 8         & 0.82         & 0.37             \\\hline
Exp 6         & CosineAnnealingLR, T\_max=25   & 0.2     & 10         & 8         & 0.82         & 0.35             \\ \bottomrule
\end{tabular}%
}
\caption{The multiple experiments run with the scheduler and hyperparameters used, along with their results}
\label{tab:different_tests}
\end{table*}

The reported results are based on the best-performing model from these six training experiments.

The proposed transformer-based model, when trained on the Skolt Sami dataset, performed well on both POS and Contlex classification tasks. The best-performing model (Exp 3) achieved an average weighted F1 score of 1.00 for POS prediction and 0.81 for Contlex classification. The hierarchical label encoding strategy and the use of BPE tokenization enabled the model to effectively handle data sparsity and morphological richness. The shared transformer layers provided an efficient way to learn the underlying linguistic structure, while the separate output heads allowed for precise classification for each task.

\subsection{POS Classification Results}

The POS classification results from the best-performing model (Exp 3) indicate exceptional performance, achieving 100\% precision, recall, and F1 score for nouns (\verb|N|) and verbs (\verb|V|). The detailed metrics are as follows:

\begin{table}[]
\resizebox{\columnwidth}{!}{%
\begin{tabular}{@{}lllll@{}}
\toprule
Label & Precision & Recall & F1-Score & N \\ \midrule
N     & 1.00      & 1.00   & 1.00     & 1520    \\\hline
V     & 1.00      & 1.00   & 1.00     & 338     \\ \bottomrule
\end{tabular}%
}
\caption{Classification results for predicting the POS using the best model}
\label{tab:pos-results}
\end{table}

The weighted average metrics for all POS labels showed perfect scores across all evaluation criteria. Specifically, the precision, recall, F1-score, and accuracy metrics were all measured at 1.00, indicating that the model correctly classified every instance without any errors for both nouns and verbs. This level of performance suggests that the model has successfully learned to distinguish between the different parts of speech in the dataset with complete reliability.

\subsection{Contlex Classification Results}

For Contlex classification, the best model (Exp 3) performed well overall, although there were notable differences in performance across various categories. The macro-averaged F1 score was 0.84, indicating that while the model performed well for many categories, some rare categories were challenging to predict accurately. Below are notable results for selected Contlex categories:

\begin{itemize}
    \item \textbf{N\_SAJOS}: Precision = 0.82, Recall = 0.83, F1-Score = 0.82 (Support = 597)
    \item \textbf{N\_MAINSTUMMUSH}: Precision = 0.33, Recall = 0.32, F1-Score = 0.33 (Support = 156)
    \item \textbf{V\_LAUKKOOLLYD}: Precision = 0.91, Recall = 0.80, F1-Score = 0.85 (Support = 61)
\end{itemize}
The detailed metrics show that for frequent categories like \verb|N_SAJOS|, the model performs well, achieving an F1 score of 0.82. However, for less frequent categories like \verb|N_MAINSTUMMUSH|, performance drops, reflecting challenges in predicting low-frequency classes. 

The precision, recall, F1-score, and accuracy for the continuation lexicon classification were all recorded at approximately 0.81, indicating that the model was able to consistently achieve a balanced level of performance across all metrics. This suggests that the model is reliable in its classification for most categories, although there is still room for improvement, particularly in handling rare classes.

These results indicate that data sparsity affects performance on less frequent labels. The comparison across different experiments further highlighted the sensitivity of model performance to hyperparameter choices, such as the number of transformer layers and dropout rates. The results from experiments 5 and 6, which achieved lower scores, underscore the importance of carefully tuning these parameters to avoid underfitting or overfitting. Data augmentation using miniparadigms helped mitigate some of these challenges, but further improvements could be achieved by expanding the dataset or incorporating additional contextual features.

\subsection{Accuracy per number of words}

\begin{figure}[t]
  \includegraphics[width=\columnwidth]{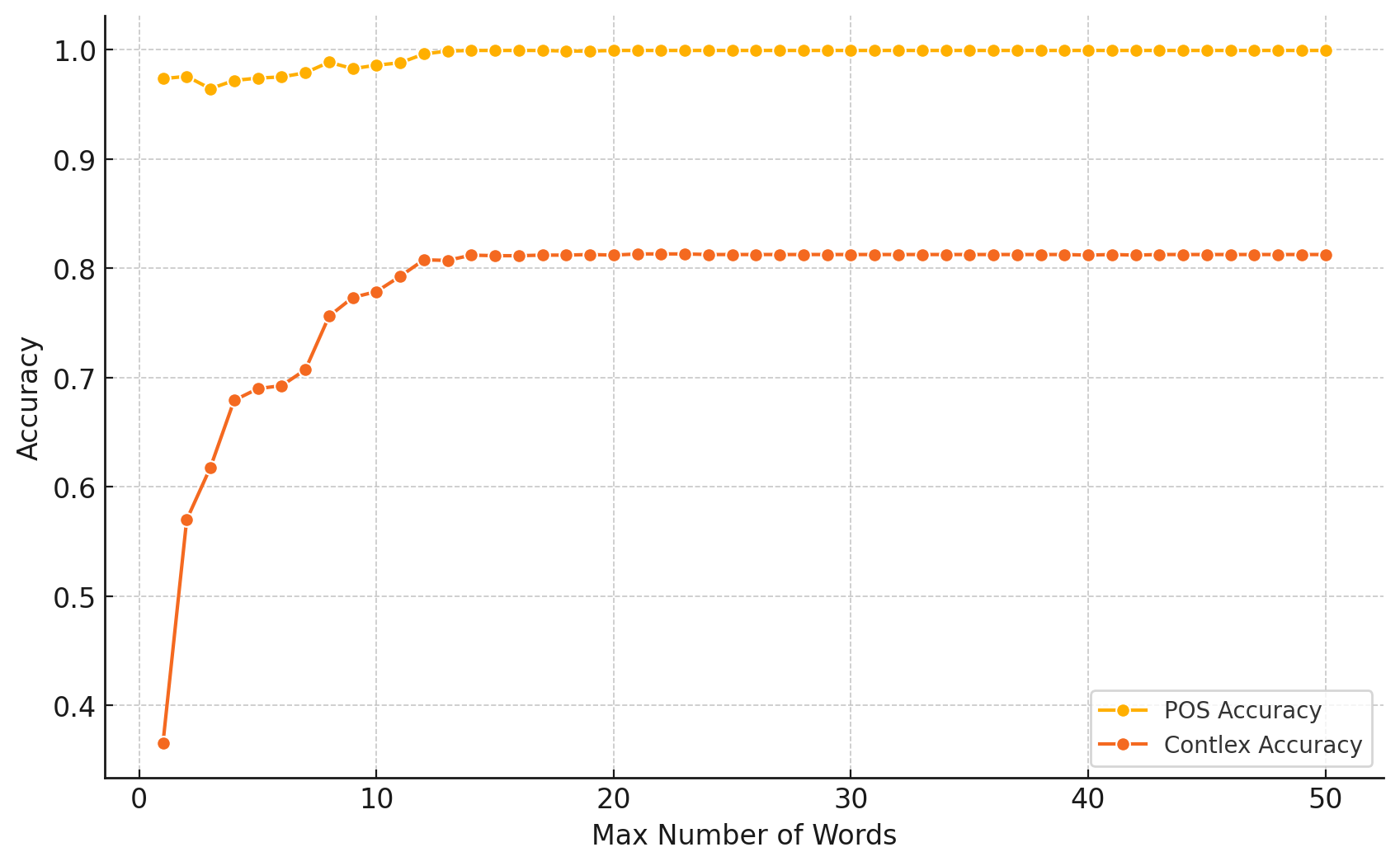}
  \caption{POS and Contlex accuracy by maximum number of word forms that are sent to the model for prediction}
  \label{fig:accuracy-per-number-of-words}
\end{figure}

We also evaluated the model's performance by limiting the maximum number of word forms sent to the model for prediction. Figure \ref{fig:accuracy-per-number-of-words} illustrates how the accuracy of POS and Contlex classification changes with an increasing number of word forms provided to the model. The results showed that both POS and Contlex accuracy improved as the number of word forms increased, eventually reaching a stable high performance. Specifically, POS accuracy started at 0.973 when the maximum number of word forms was 1 (just the lemma), and steadily improved, reaching 0.999 for 14 or more word forms. Similarly, Contlex accuracy improved from 0.365 at 1 word form to 0.69 for 5 word forms and to above 0.81 for 14 or more word forms. This demonstrates that providing more paradigmatic context significantly enhances the model's ability to make accurate predictions.

\section{Discussion and Conclusion}

In this paper, we presented a transformer-based approach for predicting parts of speech and inflection classes (Contlexes) for the Skolt Sami language. The success of the model highlights the potential of combining traditional linguistic tools with modern NLP techniques, particularly for endangered languages. Our results demonstrate near-perfect performance for POS classification and reasonably good performance for most Contlex categories, although predicting rare categories remains challenging. The results indicate that the use of shared embeddings and multi-task learning can be effective in achieving high accuracy for parts of speech, while data augmentation and careful hyperparameter tuning help in handling the morphological complexities of Skolt Sami.

The observed variability in Contlex classification performance, especially for infrequent categories, highlights the challenges of data sparsity and suggests the need for additional efforts in data collection and augmentation. Frequent categories like \verb|N_SAJOS| benefited from the availability of more examples, whereas rare categories such as \verb|N_MAINSTUMMUSH| showed lower performance, primarily due to limited training data. This underscores the necessity for expanding the training dataset to cover more diverse lexical entries and reduce biases towards common categories. Incorporating additional features, such as syntactic or contextual information, could also enhance the model’s understanding of rare categories.

The results from limiting the number of words used for prediction suggest that context plays a crucial role in improving model performance. When fewer words were provided to the model, both POS and Contlex accuracy suffered, indicating the importance of sufficient contextual information for effective classification. The model showed a consistent improvement in both tasks as more words were added, and the performance eventually stabilized. This demonstrates that using larger contexts allows the transformer model to better capture the linguistic intricacies of Skolt Sami, improving the reliability of its predictions.

Moreover, we believe that expanding the dataset to include other related Uralic languages could enhance model performance through cross-linguistic transfer learning, benefiting from shared morphological features. Another promising direction for future work is the exploration of semi-supervised or unsupervised learning techniques, which could leverage unlabeled data to improve classification performance without relying solely on manually annotated resources. This is particularly relevant given the resource constraints typical for endangered languages like Skolt Sami.

In conclusion, the trained model and code will be released publicly to support future research and application in endangered language processing. We hope that this contribution will aid in the ongoing efforts to preserve and revitalize minority languages by providing computational tools that can be used to automate linguistic analysis, document new lexical entries, and contribute to the development of educational and linguistic resources. Future research should continue to focus on enriching the dataset, exploring multi-lingual training, and employing innovative learning paradigms to further advance the field of NLP for endangered languages.



\bibliography{acl_latex}



\end{document}